\documentclass[journal]{IEEEtran}
\newtheorem{definition}{Definition}

\newenvironment{proof}{\begin{IEEEproof}}{\end{IEEEproof}}
\usepackage{amssymb}
\newtheorem{strategy}{Strategy}    
\newtheorem{property}{Property}
\usepackage{multirow}  
\usepackage{hyperref}

\usepackage{amsmath}

\usepackage[ruled,linesnumbered]{algorithm2e}

\ifCLASSINFOpdf
\usepackage[pdftex]{graphicx}
\DeclareGraphicsExtensions{.pdf,.jpeg,.png}
\else
\usepackage[dvips]{graphicx}
\DeclareGraphicsExtensions{.eps}
\fi

\ifCLASSINFOpdf
\else
\fi

\begin{document}

\title{Towards Correlated Sequential Rules}

\author{Lili Chen, Wensheng Gan*, Chien-Ming Chen*,~\IEEEmembership{Senior Member,~IEEE}

\thanks{This research was supported in part by the National Natural Science Foundation of China (Grant Nos. 62002136 and 62272196), Natural Science Foundation of Guangdong Province (Grant No. 2022A1515011861), Guangzhou Basic and Applied Basic Research Foundation (Grant No. 202102020277), and the Young Scholar Program of Pazhou Lab (Grant No. PZL2021KF0023).}
	
	\thanks{Lili Chen is with the Department of Computer Science and Technology, Tongji University, Shanghai 200082, China. E-mail: lilichien3@gmail.com} 
	
	\thanks{Wensheng Gan is with the College of Cyber Security, Jinan University, Guangzhou 510632, China; and with Pazhou Lab, Guangzhou 510330, China. E-mail: wsgan001@gmail.com}
	
	\thanks{Chien-Ming Chen is with the College of Computer Science and Engineering, Shandong University of Science and Technology, Qingdao 266590, China. E-mail: chienmingchen@ieee.org} 
	
	\thanks{Corresponding author: Wensheng Gan and Chien-Ming Chen}
}


\maketitle


\begin{abstract}
 The goal of high-utility sequential pattern mining (HUSPM) is to efficiently discover profitable or useful sequential patterns in a large number of sequences. However, simply being aware of utility-eligible patterns is insufficient for making predictions. To compensate for this deficiency, high-utility sequential rule mining (HUSRM) is designed to explore the confidence or probability of predicting the occurrence of consequence sequential patterns based on the appearance of premise sequential patterns. It has numerous applications, such as product recommendation and weather prediction. However, the existing algorithm, known as HUSRM, is limited to extracting all eligible rules while neglecting the correlation between the generated sequential rules. To address this issue, we propose a novel algorithm called correlated high-utility sequential rule miner (CoUSR) to integrate the concept of correlation into HUSRM. The proposed algorithm requires not only that each rule be correlated but also that the patterns in the antecedent and consequent of the high-utility sequential rule be correlated. The algorithm adopts a utility-list structure to avoid multiple database scans. Additionally, several pruning strategies are used to improve the algorithm's efficiency and performance. Based on several real-world datasets, subsequent experiments demonstrated that CoUSR is effective and efficient in terms of operation time and memory consumption.
 \\
 \textit{Impact Statement} -- This article contributes to a correlation-based high-utility sequential rule discovery model for data prediction and artificial intelligence analytics. To the best of our knowledge, it is the first article that proposes a realistic correlation-based solution for high-utility sequential rule mining instead of discovering patterns or rules uncorrelated in real-world datasets. The successful application of rule discovery and high-utility pattern mining algorithms can bring great business value in engineering management and profit generation. Since CoUSR can accurately predict the occurrence of sequential patterns with correlation, it can be used in many different applications and domains, such as market basket analysis, risk prediction, and intrusion detection.
\end{abstract}



\IEEEpeerreviewmaketitle

\section{Introduction}

\IEEEPARstart{A} wide variety of devices generate massive amounts of data, precipitating the arrival of the data age. Data mining has become a popular research area because it uncovers useful and valid information in vast amounts of data, promoting economic and social progress \cite{feelders2002data, chen1996data}. Initially, researchers considered that items in transaction databases either existed or did not, and that the extracted patterns occurred frequently, which is called frequent pattern mining (FPM) \cite{agrawal1994fast, han2000mining}. The FPM assumes that all items in a transaction occur simultaneously, but in reality, items occur in a specific order. A sequential pattern is a collection of items with a time dimension, and sequential pattern mining (SPM) is the process of discovering critical information derived from sequential databases \cite{fournier2017survey, gan2019survey, han2000freespan,wu2020netncsp}. SPM is expected to find sequential patterns that satisfy the support requirements. Its application involves a host of aspects, such as web logs and market analysis. Agrawal and Srikant \cite{srikant1996mining} first introduced the problem of SPM in the early 1990s. However, even though some frequently occurring sequential patterns are known, there is insufficient evidence to make predictions regarding the events. For example, suppose there is a frequent sequential pattern $ab$ in which $a$ may occur before or after $b$, it is unwise to apply this pattern to predict whether $a$ occurs and $b$ follows.

In terms of chronological properties and prediction of events, sequential rule mining (SRM) \cite{fournier2014erminer, fournier2012cmrules} outperforms SPM. A sequential rule states that if some items occur as antecedents, some non-overlapping items may appear later as consequent with a given degree of confidence. The task of mining rules aims to predict patterns or events that have not yet occurred. We can predict the likelihood or relevance of a pattern in the future based on the existence of a pattern. Several algorithms have been proposed for SRM. Some of them discover rules from a single sequence, and others find rules from multiple sequences. Some of the sequential rules, both the antecedent and the consequent, are sequential patterns, while the others are composed of two unordered collections of items. Mannila \textit{et al.} \cite{mannila1997discovery} discovered frequent events partially ordered in a sequence. Hamilton \textit{et al.} \cite{hamilton2005timers} presented a temporal investigation method for registered record sequences. Harms \textit{et al.} \cite{harms2002discovering} proposed the MOWCATL algorithm to find rules whose support and confidence are both greater than the corresponding thresholds. In this study, we only consider SRM from a few sequences and partially disordered items.

In terms of FPM, it removes the limitations that items in a transaction may appear more than once and that the weight or profit of items varies in real life \cite{geng2006interestingness}. Therefore, FPM leads to the discovery of frequent patterns, but the profits obtained are not satisfactory, particularly for dissatisfied retailers or manufacturers. To address these issues, high-utility pattern mining (HUPM) \cite{gan2021survey} was developed, which considers not only the number of patterns but also their unit profit. In general, HUPM can be divided into two categories: two-phase and one-phase algorithms, where the former generates many candidates, but the latter does not. Some representative two-phase algorithms include Two-Phase \cite{liu2005two}, UP-Growth \cite{tseng2010up}, and UP-Growth$^+$ \cite{tseng2012efficient}. Representative one-phase algorithms include HUI-Miner \cite{liu2012mining} and FHM \cite{fournier2014fhm}. High-utility sequential pattern mining (HUSPM) is derived by extending HUPM to the temporal aspect \cite{yin2012uspan, yin2013efficiently}. Yin \textit{et al.} \cite{yin2012uspan} developed a generic framework and proposed an efficient algorithm called USpan. They designed a lexicographic quantitative sequence tree for searching full high-utility sequential patterns (HUSPs). To cope with the requirements of prediction with utility, HUSRM \cite{zida2015efficient} was developed. HUSRM \cite{zida2015efficient} is a one-phase algorithm that depends on the utility-table structure. Additionally, several optimization strategies have been employed to improve the performance of the algorithm.

However, because these rules are constrained only by utility and confidence, the items in those are uncorrelated, which may lead to decision-making errors \cite{gan2019correlated, fournier2020mining}. For example, the existing algorithm reveals that the rule of first selling a TV and bread together in a sequential database, followed by apples and sofas being sold together with a certain degree of confidence, is a HUSR. It is clear that the items in the antecedent and the consequent of the rule are hardly correlated, and the rule’s antecedent and consequent are also weakly correlated. To date, several correlated HUPM algorithms have been developed, such as HUIPM \cite{ahmed2011framework}, FDHUP \cite{lin2017fdhup}, and CoUPM \cite{gan2019correlated}. However, the algorithms for mining no correlated HUSRs (CHUSRs) have been proposed.

To deal with the problems mentioned above, we incorporate the concept of correlation to HUSRM. Here are the new ideas and contributions of this study:

\begin{itemize}
	\item This study develops an efficient CoUSR algorithm for mining CHUSRs from a given sequence database. To the best of our knowledge, this is the first study to involve mining correlated utility-based sequential rules.
	
	\item The correlation of rules has two aspects. On the one hand, it is a local correlation that the antecedent or consequent within the rule is constrained by the \textit{bond} measure. On the other hand, it is a global correlation in which sequential rules are constrained by the \textit{lift} method.
	
	\item CoUSR is a one-phase algorithm that uses a utility-list structure to maintain data and does not require multiple database scans. Two novel structures called BondMatrix and estimated sequence utility co-occurrence structure (ESUCS) are constructed, and several pruning strategies are designed to assist in pruning those unpromising sequential rules and reduce the search space.
	
	\item To evaluate the performance of the algorithm, subsequent experiments measure the efficiency and feasibility of the proposed algorithm under different strategies.
\end{itemize}

The remainder of this paper is organized as follows. Section \ref{sec:Related} presents relevant literature on the proposed algorithm. Some preliminaries and problem statements are introduced in Section \ref{sec:background}. Section \ref{sec:Algorithm} describes the details of the designed strategies and the proposed algorithm. The experimental comparison is presented in Section \ref{sec:experiments}. Finally, Section \ref{sec:conclusion} concludes this study and highlights future studies.

\section{Related Work}  \label{sec:Related}

In this section, we review the related literature on the proposed algorithm, including HUSPM, SRM, and CHUPM.

\subsection{High-Utility Sequential Pattern Mining}

Frequency-based sequential pattern mining \cite{fournier2017survey,gan2019survey} has attracted a lot of attention in the past decades. High-utility SPM (HUSPM for short) plays an important role in decision-making. To date, a series of relevant algorithms have been developed. The SPM was originally proposed by Agrawal and Srikant \cite{agrawal1995mining} to address the challenge of discovering meaningful subsequences in a collection of sequences. In terms of utility-based issues, Ahmed \textit{et al.} \cite{ahmed2010novel} introduced two algorithms in 2010: UtilityLevel, which applies the method of candidate generation and testing, and UtilitySpan, which uses a pattern growth approach. Yin \textit{et al.} \cite{yin2012uspan} then developed a generic framework called USpan to address the HUSPM problem. A lexicographic quantitative sequence tree and two concatenation mechanisms were also proposed \cite{yin2012uspan}. To improve the efficiency, several advanced algorithms with new data structures and new pruning strategies, such as HUS-Span \cite{wang2016efficiently}, ProUM \cite{gan2020proum}, HUSP-ULL \cite{gan2021fast} have been proposed. In addition to efficiency, the effectiveness is also the key issue of data mining algorithms, thus some studies focus on mining top-$k$ HUSPs \cite{yin2013efficiently,wang2016efficiently,zhang2021tkus}, incremental HUSPM \cite{wang2018incremental}, HUSPM over data streams \cite{zihayat2017memory}, HUSPM with negative item values \cite{xu2017mining}, mining multi-dimensional HUSPs using MDUS$_{EM}$ and MDUS$_{SD}$  \cite{gan2021utility}, and on-shelf availability for HUSPM  \cite{zhang2021shelf}. Additionally, MAHUSP \cite{zihayat2017memory} employed memory-adaptive mechanisms by taking advantage of the upper bound of memory, and the MAS-tree was also designed to maintain potential HUSPs. Other related works of utility-driven SPM and advanced utility mining tasks can be referred to the review literature \cite{gan2021survey}.


\subsection{Sequential Rule Mining}

Different from sequential pattern mining (SPM) \cite{gan2019survey,van2018mining}, the SRM uses rules from frequent patterns in a set of sequences, which works well in prediction. Lo \textit{et al.} \cite{lo2009non} considered that a rule is redundant when it has the same level of support and confidence as inferred rules. Based on logical inference, a non-redundant rule was characterized, and a related algorithm was proposed. Then, Fournier-Viger \textit{et al.} \cite{fournier2012cmrules} proposed CMRules to generalize the point that there is no order for items in the antecedent and consequent of all rules. Its intention is to first transform the sequence database into a transaction database by ignoring the temporal information, then mine association rules in the transaction database, and then filter out sequential rules in these association rules through the original sequence database. Obviously, this would be inefficient. Subsequently, RuleGrowth \cite{fournier2011rulegrowth} was proposed using a pattern-growth approach, which is more efficient and scalable. First, the shortest rules are found, and then the left and right expansions are used to find full sequential rules.  
TRuleGrowth \cite{fournier2015mining} was proposed, and it created a sliding window to manage the maximum number of rule occurrences. The utility-driven SRM is introduced for utility-based sequential rule mining by extending sequential rule mining for utilities. To date, there are only two relevant algorithms: HUSRM \cite{zida2015efficient} and e-HUNSR \cite{zhang2020hunsr}. The HUSRM combines sequential rules with utility to establish a framework of utility-based sequential rule mining. It is primarily depth-first searching rules, and the mining procedure is similar to RuleGrowth \cite{fournier2011rulegrowth}. The HUSRM algorithm employs sequence-estimated utility as an upper bound of the utility of sequential rules, which can trim off some unpromising items and rules in advance. Furthermore, using the information in the utility-list structure not only reduces the database access but also accelerates the mining process by revealing a tighter upper bound on utility. On this basis, e-HUNSR \cite{zhang2020hunsr} considered that some events do not occur. e-HUNSR proposed some solutions to address the challenges of some intrinsic complexities. 

\subsection{Correlated High-Utility Pattern Mining}

Typical HUPM algorithms use utility as the sole criterion, resulting in almost no correlation between items in the patterns. To prevent this situation, a number of algorithms have been proposed to consider both utility and correlation to discover correlated high-utility patterns (CHUPs). The HUIPM was originally proposed to mine high-utility interesting patterns with frequency affinity \cite{ahmed2011framework}. In this algorithm, the knowledge-weighted utilization (KWU) of a pattern is used as an upper bound to prune patterns in advance. Then, FDHUP primarily improved the efficiency of HUIPM by developing a novel EI-table and FU-table data structure \cite{lin2017fdhup}. Following that, FCHM-bond and FCHMall-confidence were proposed as bond and all-confidence measures \cite{fournier2020mining,fournier2016mining}, respectively. Both of these methods have anti-monotonicity, which makes pruning patterns easier. Subsequently, CoHUIM \cite{gan2018extracting}, CoUPM \cite{gan2019correlated}, and CoHUI-Miner \cite{vo2020mining} were developed to discover high-utility patterns using the Kulc measure \cite{geng2006interestingness}. They are all one-phase algorithms that employ projection databases and utility lists to store significant information. Clearly, the concept of correlation is important in real-world applications. The problem of mining CHUSRs is discussed in this study. Our CoUSR algorithm discovers pattern information with sequential properties on the basis of CoUPM \cite{gan2019correlated} for more valuable patterns, which is an extension of the former one.

\section{Preliminary and Problem Statement} 
\label{sec:background}

To better explain the proposed algorithm, some essential definitions are introduced in advance. A sequential database \textit{SD} consists of multiple sequences denoted as \textit{SD} = \{\textit{S}$_{1}$, \textit{S}$_{2}$, $\ldots$, \textit{S$_{n}$}\}, where 1, 2, $\ldots$, and $n$ represent sequence identifiers. Let $I$ = \{\textit{i}$_{1}$, \textit{i}$_{2}$, $\ldots$, and \textit{i$_{m}$}\} be the items in the sequential database. A sequence $S$ is composed of a collection of itemsets with temporal order $\prec$, which is expressed as $S$ = \{\textit{I}$_{1}$, \textit{I}$_{2}$, $\ldots$, \textit{I$_{l}$}\}, where \textit{I$_{l}$} is not only a subset of $I$ but also disjoint subsets of $I$. Additionally, because there may be more than one item in a sequence $S$, $q(i, S)$ is employed to indicate the number of items $i$ in $S$, that is, its internal utility. Each item is associated with a value $p(i)$ corresponding to the weight (i.e., external utility). In a sequence, there is an order between itemsets and a disorder within them. To illustrate examples, a sequence database and a utility table are shown in Tables \ref{table:SD} and \ref{table:profit}, respectively.

\begin{table}[h]
	\centering
	\small
	\caption{A sequence database.}
	\label{table:SD}
	\begin{tabular}{|c|c|c|c|}
		\hline
		\textbf{\textit{SID}} & \textbf{Sequence} & \textbf{\textit{SEU}} \\ \hline 
		$ S_{1} $ & 	\{\textit{a}:1, \textit{b}:1\}, \textit{e}:1, \textit{d}:5, \textit{g}:1  &  \$21 \\ \hline
		$ S_{2} $ & 	\{\textit{a}:2, \textit{d}:9\}, \textit{c}:2, \textit{b}:1, \{\textit{e}:1, \textit{g}:2\}  & \$34 \\ \hline
		$ S_{3} $ &	    \textit{a}:1, \textit{b}:2, \textit{f}:1, \textit{e}:2  &  \$28 \\ \hline
		$ S_{4} $ &	    \{\textit{a}:1, \textit{b}:1, \textit{d}:2\}, \textit{e}:1, \textit{g}:3  &  \$22 \\ \hline
		$ S_{5} $ &	    \{\textit{a}:3, \textit{b}:1\}, \textit{e}:1, \textit{f}:3, \textit{c}:4, \textit{d}:3, \textit{g}:1 &   \$42  \\ \hline
	\end{tabular}
\end{table}

\begin{table}[h]
	\centering
	\small
	\caption{Unit utility of each item}
	\label{table:profit}
	\begin{tabular}{|c|c|c|c|c|c|c|c|}
		\hline
		\textbf{Item}    &  $ a $  &  $ b $  &  $ c $   & $ d $  & $ e $  & $ f $ & $ g $\\ \hline
		\textbf{Utility} &  $ 3 $  &  $ 5 $  &  $ 2 $   & $ 1 $  & $ 6 $  & $ 3 $ & $ 2 $\\ \hline
	\end{tabular}
\end{table}

As shown in Table \ref{table:SD}, there are five sequences, $S_1$, $S_2$, $S_3$, $S_4$, and $S_5$ in the sequence database. There are four itemsets in $S_1$, which are \{$a$, $b$\}, $e$, $d$, and $g$ in chronological order. Among them, the quantity or internal utility of $a$ is 1, and that of $b$ is one. Additionally, as shown in Table \ref{table:profit}, the external utilities of $a$ to $g$ are \$3, \$5, \$2, \$1, \$6, \$3, and \$2.

\begin{definition}
	\rm Given two unordered non-empty itemsets $X$ and $Y$, they can form a rule $r$: $X \Rightarrow Y$ if $X$ and $Y$ are subsets of $I$ and there is no intersection between them. Rule $r$ indicates that if $X$ appears in a sequence, then $Y$ will also appear in the same sequence. Assume the number of items in $X$ is $k$ and the number of items in $Y$ is $m$, implying that $|X|$ = $k$ and $|Y|$ = $m$. Then, the size of a rule $r$: $X \Rightarrow Y$ is defined as $k \ast m$, where the symbol $\ast$ does not represent a product. Let the size of another rule $r'$ be $t \ast u$. If $t > k$ and $u \geq m$, or $t \geq k$ and $u > m$, we consider $r'$ to be greater than $r$ naturally.
\end{definition}

Consider a simple example: the size of $r'$: \{$a$, $b$, $c$, $d$\} $\Rightarrow$ \{$e, f$\} is 4 $\ast$ 2 and the size of $r$: $\{a\}$ $\Rightarrow$ $\{f\}$ is 1 $\ast$ 1. Because 4 $>$ 2 and 2 $>$ 1, we can come to the conclusion that the size of $r'$ is greater than that of $r$.

\begin{definition}
	\rm Let a sequence $S_c$ = \{\textit{I}$_{1}$, \textit{I}$_{2}$, $\ldots$, \textit{I$_{l}$}\} contain $l$ itemsets. If the itemset \textit{I$_{x}$} is a union of some itemsets in $S_c$, \textit{I$_{x}$} occurs in $S_c$.  For a rule $r$: $X \Rightarrow Y$, assuming that there exists an integer $p$, such that $ l >$ $p \geq$ 1, $X$ $\subseteq$ $\cup_{i = 1}^{p} I_i$ and $Y$ $\subseteq$ $\cup_{i = p + 1}^{l} I_i$, we consider that $r$ appears in $S_c$ or $S_c$ supports $r$. We denote the sequences containing a sequential rule $r$ as \textit{sids}$(r)$, and the sequences comprising an itemset $X$ are denoted as \textit{sids}$(X)$. The \textit{confidence} of a rule $r$: $X$ $\Rightarrow$ $Y$ is defined as \textit{conf}$(r)$ = $|\textit{sids}(r)|$/$|\textit{sids}(X)|$.
\end{definition}

As shown in Table \ref{table:SD}, the rule $a \Rightarrow b$ occurs in sequences $S_2$ and $S_3$; therefore, \textit{sids}$(r)$ = \{$S_2$, $S_3$\}. Similarly, $a$, the antecedent of $r$ appears in all five sequences, that is, \textit{sids}$(a)$ = \{$S_1$, $S_2$, $S_3$, $S_4$, $S_5$\}. Therefore, the confidence level of $r$ is calculated to be 2/5 = 0.4.

\begin{definition}
	\rm The utility of an item $i$ in a sequence $S_c$ is denoted as $u(i, S_c)$ = $q(i, S_c)$ $\times$ $p(i)$. The utility of a rule $r$: $X$ $\Rightarrow$ $Y$ in $S_c$ is defined as $u(r, S_c)$ = $\sum_{i \in X}u(i, S_c)$ + $\sum_{i \in Y}u(i, S_c)$ when $r$ occurs in $S_c$. Moreover, we define the utility of $r$ in the sequence database \textit{SD} as the sum of utilities of $r$ in all sequences containing it, and is denoted as $u(r)$ = $\sum_{S_c \in SD}u(r, S_c)$ \cite{zida2015efficient}.
\end{definition}

Following the above example, the utilities of items $a$ and $b$ in the sequence $S_2$ can be calculated as 2 $\times$ \$3 = \$6 and 1 $\times$ \$5 = \$5, respectively. Thus, the utility of $r$ is equal to the sum of $a$ and $b$ utilities in $S_2$, which is \$11. Furthermore, the utility of $r$ in the sequence database is calculated as $u(r, S_2)$ + $u(r, S_3)$ = \$24.

\begin{definition}
	A high-utility sequential rule (HUSR) $r$ must satisfy two conditions: its confidence must be greater than or equal to the minimum confidence threshold \textit{minconf} $\in$ $[0, 1]$, and its utility must be greater than or equal to the minimum utility threshold \textit{minutil} $\in$ $R^+$, which can be denoted as \textit{conf}$(r)$ $\geq$ \textit{minconf} and $u(r) \geq$ \textit{minutil}, respectively. The task of discovering HUSRs from a sequence database is to determine all valid rules.
\end{definition}

Although HUSRs have a broad application in prediction, there are some disadvantages, such as the fact that the obtained rules are weakly correlated, which causes some decision-making issues. To address this issue, we present two types of correlation concepts, called \textit{bond} and \textit{lift}, and incorporate them into HUSRs to form the concept of CHUSRs.

\begin{definition}
	\label{bond}
	\rm We define the \textit{support} of a rule $r$ denoted as \textit{sup}$(r)$ = $|$\textit{sids(r)}$|$ in a sequence database. Moreover, the \textit{support} of an itemset $X$ is defined as \textit{sup}$(X)$ = $|$\textit{sids(X)}$|$. The \textit{disjunctive support} of an itemset $X$ is defined as the number of transactions containing any item in $X$ and is denoted as \textit{dissup}$(X)$ = $|\{S_c \in SD$ $| X \cap$ $S_c \neq$ $\emptyset\}|$. The \textit{bond} of an itemset $X$ is denoted as \textit{bond}$(X)$ = \textit{sup}$(X)$ / \textit{dissup}$(X)$.
\end{definition}

\begin{definition}
	\rm A rule $r$: $X \Rightarrow Y$ is locally correlated if and only if its antecedent and consequent satisfy that \textit{bond}$(X)$ $\geq$ \textit{minbond} and \textit{bond}$(Y)$ $\geq$ \textit{minbond}, where \textit{minbond} $ \in$ $[0, 1]$ is a predetermined expert-specified minimum bond threshold.
\end{definition}

For rule $r$, $\{a, b\} \Rightarrow \{g\}$ appearing in sequence $S_1$, $S_2$, $S_4$, and $S_5$, \textit{sup}$(r)$ is calculated as four. Because itemset $\{a, b\}$ occurs in five sequences, \textit{sup}$(\{a, b\})$ is five. Because each sequence contains a subset of $\{a, b\}$, \textit{dissup}$(\{a, b\})$ is also equal to five. Therefore, \textit{bond}$(\{a, b\})$ is calculated as \textit{sup}$(\{a, b\})$/\textit{dissup}$(\{a, b\})$ = 1. Because the bond of a single item is always one, \textit{bond}$(\{g\})$ = 1. If \textit{minbond} is 0.3, $r$ is locally correlated.

\begin{definition}
	\rm Given a rule $r$: $X \Rightarrow Y$, the \textit{lift} of $r$ is defined as \textit{lift}$(r)$ = \textit{conf}$(r)$ / (\textit{sup}$(Y)$/|SD|), simplified to ($|SD|$ $\times$ \textit{sup}$(r))$ / ($\textit{sup}$($X$) $\times$ \textit{sup}$(Y)) $ \cite{geng2006interestingness}. A rule $r$: $X \Rightarrow Y$ is globally correlated if it satisfies the condition that \textit{lift}$(r)$ $\geq$ \textit{minlift}, where \textit{minlift} $ >$ 1 is a predefined minimum lift threshold. If a HUSR is both locally and globally correlated, we regard it as a CHUSR.
\end{definition}

\textit{lift}$(r)$ denotes the ratio of the support of $Y$ under the condition of $X$ and that of $Y$ without any condition \cite{geng2006interestingness}. If \textit{lift}$(r)$ = 1, it indicates that $X$ and $Y$ are independent of each other and $X$ has no lifting effect on the occurrence of $Y$. A larger value indicates a greater lift of $X$ on $Y$ and a stronger correlation between them. If the value of \textit{lift}$(r)$ is less than one, the antecedent and consequent of the rule are mutually exclusive. If the value is one, the two components of $r$ are independent. If the value is greater than one, the rule is strongly correlated. To ensure that the derived rules are correlated, we established a lift range greater than one. In the running example, \textit{lift}$(\{a, b\}$ $\Rightarrow$ $\{g\})$ = (5 $\times$ 4) / (5 $\times $ 4) = 1, which is equal to one. Therefore, this rule is not globally correlated. When we set \textit{minconf} to 0.7, \textit{minutil} to 50, \textit{minbond} to 0.3, and \textit{minlift} to 1.1, all the generated rules are listed in Table \ref{table:rules}.

\begin{table}[h]
	\centering
	\small
	\caption{Desired correlated high-utility sequential rules}
	\label{table:rules}
	\begin{tabular}{|c|c|c|c|}
		\hline
		\textbf{CHUSR}    &  \textbf{\textit{utility}}  &  \textbf{\textit{conf}}  &   \textbf{lift}     \\ \hline \hline
		$\{a, b, c, d\} \Rightarrow \{g\}$ &  $ 55 $  &  $ 1.0 $  & 1.25  \\ \hline
		$\{a, b, d\} \Rightarrow \{g\}$ &  $ 74 $  &  $ 1.0 $  &  1.25  \\ \hline
		$\{a, d\} \Rightarrow \{g\}$ &  $ 54 $  &  $ 1.0 $  &  1.25   \\ \hline
		$\{b, d\} \Rightarrow \{g\}$ &  $ 53 $  &  $ 1.0 $  &  1.25  \\ \hline
	\end{tabular}
\end{table}

\section{Proposed CoUSR Algorithm} \label{sec:Algorithm}

In this section, we first introduce the downward closure properties of utility and correlation, and then present a compact data structure named utility-list to store critical information. Subsequently, some pruning strategies are proposed to optimize the performance of the CoUSR. Finally, the details and pseudocode of the proposed algorithm are presented.

\subsection{Anti-monotonicity of Sequence Estimated Utility and the Bond Measure}

In previous studies, it was discovered that utility lacks monotonicity and anti-monotonicity, contributing to an explosion in the search space. To address this dilemma, the sequence-estimated utility is proposed to prune sequential rules.

\begin{definition}
	\label{def_12}
	\rm We define the \textit{sequence utility} as the total of utilities for each item in a sequence $S_c$, which can be denoted as $SU(S_c)$ = $\sum_{i \in S_c}u(i, S_c)$. Additionally, the sequence-estimated utility of an item $i$ is defined as the sum of sequence utilities where sequences contain $i$. The expression can be written as \textit{SEU}$(i)$ = $\sum_{i \in S_c \wedge S_c \in SD}SU(S_c)$. We define the sequence-estimated utility of a rule $r$ as the summary of the sequence utility of the sequences containing rule $r$, and write it as \textit{SEU}$(r)$ = $\sum_{ S_c \in \textit{sids}(r)}SU(S_c)$ \cite{zida2015efficient}.
\end{definition}

The last column in Table \ref{table:SD} shows the sequence utility of each sequence. According to the calculation results, $SU(S_1)$ = \$21, $SU(S_2)$ = \$34, $SU(S_3)$ = \$28, $SU(S_4)$ = \$22, and $SU(S_5)$ = \$42. It is already known that the rule $r$: $X \Rightarrow Y$ appears in the sequence $S_2$ and $S_3$; thus, \textit{SEU}$(r)$ = $SU(S_2)$ + $SU(S_3)$ = \$62.

According to Definition \ref{def_12}, the \textit{SEU} of an item or rule must be no less than its real utility, as well as the utility of its expansion rules. Hence, \textit{SEU} serves as an upper bound on the utility. It is apparent that \textit{SEU} has a downward closure property, which grounds the two properties introduced below.

\begin{property}
	\label{pro_1}
	\rm For an item $i$, assuming that \textit{SEU}$(i) <$ \textit{minutil}, the item is said to be unpromising; otherwise, it is promising. Furthermore, if an item is unpromising, the rules containing that item are unlikely to be eligible \cite{zida2015efficient}.
\end{property}

\begin{property}
	\label{pro_2}
	\rm For a rule $r$, if \textit{SEU}$(r) < $ \textit{minutil}, $r$ is believed to be unpromising; otherwise, $r$ is promising. Neither are the rules derived from the expansion of this rule, which is expected to be promising \cite{zida2015efficient}.
\end{property}

Based on these two properties, a few pruning strategies can be used to remove some items and rules that are unsatisfactory.

\begin{strategy}
	\label{stra_1}
	\rm If the inequality \textit{SEU}$(i) < $ \textit{minutil} for an item $i$ holds, this item and all sequential rules containing $i$ can be pruned directly.
\end{strategy}

\begin{strategy}
	\label{stra_2}
	\rm If the inequality \textit{SEU}$(r) < $ \textit{minutil} for a rule $r$ holds, $r$ and its expansions can be pruned directly.
\end{strategy}

According to Definition \ref{bond}, the bond measure has a downward closure property because, as the number of items in itemset $X$ increases, the number of sequences containing $X$ decreases, while the number of sequences containing any item in $X$ increases, decreasing the bond of $X$.

\begin{property}
	\label{pro_bond}
	\rm For a rule $r$: $X \Rightarrow Y$, assuming that \textit{bond}$(X) <$ \textit{minbond} or \textit{bond}$(Y) < $ \textit{minbond}, this rule and all its expansions are unpromising.
\end{property}

We can propose the following pruning strategy based on this property.

\begin{strategy}
	\label{stra_bond}
	\rm If the bond of the antecedent or consequent of a sequential rule is less than \textit{minbond}, then the rule and any of its expansions can be pruned directly.
\end{strategy}

\subsection{Utility-list Structure}

\begin{figure}[h]
	\centering
	\includegraphics[scale=0.4]{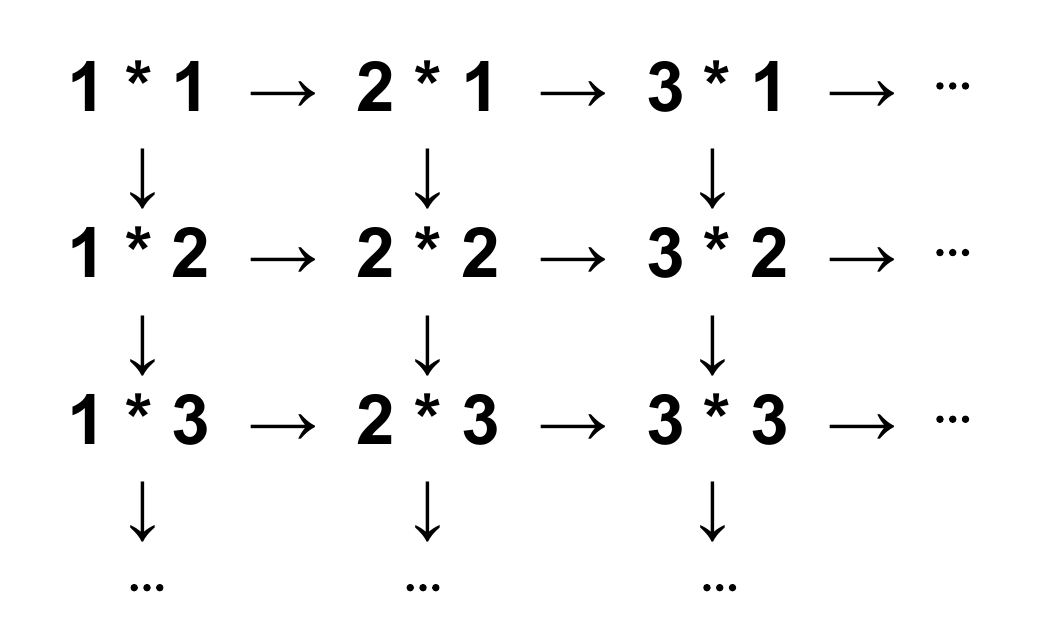}
	\caption{Unordered left-expansion and right-expansion}
	\label{fig:order1}
\end{figure}

\begin{figure}[h]
	\centering
	\includegraphics[scale=0.4]{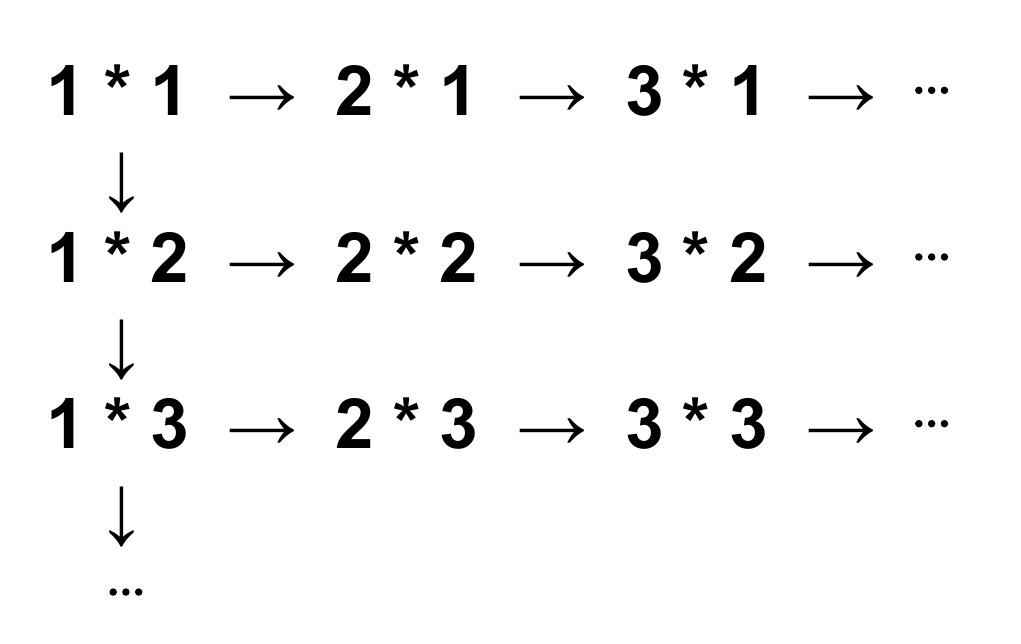}
	\caption{The order of not allowing performing a right expansion after a left expansion}
	\label{fig:order2}
\end{figure}

This subsection describes in detail a novel utility-list structure \cite{zida2015efficient}. Before that, we must address two issues that may result in redundant rules. We observe that by varying the combinations of left- and right-expand operations, we can obtain a specific rule. For example, a rule $r$: $\{a, c\} \Rightarrow \{e, g\}$ is given. If the initial rule is $\{a\} \Rightarrow \{e\}$, the target rules can be derived not only from the primitive rule by a left expansion followed by a right expansion, but also by a right expansion followed by a left expansion. As shown in Fig. \ref{fig:order1}, different trails can lead to the same results. A simple and naive approach is to allow left expansions after right expansions but not right-expansions after left expansions. According to Fig. \ref{fig:order2}, there are no redundant rules to generate. Of course, it is also possible to allow right-expansions after left-expansions, but not to permit left-expansions after right-expansions.

Another discovery potentially influencing the accuracy of the results is that it is possible to obtain the same rule for either left- or right-expansions with various items. Take an example of the rule $r$: $\{a, c\}$ $\Rightarrow$ $\{g\}$, which can be achieved either by $\{a\}$ $\Rightarrow$ $\{g\}$ through left-expansions with $c$, or by $\{c\}$ $\Rightarrow $ $\{g\}$ through left-expansion with $a$. To accommodate such problems, we restrict the expansion according to the order of the alphabet denoted by $\prec$, and in the above example, there is no possibility of expanding $\{c\}$ $\Rightarrow $ $\{g\}$ to $\{a, c\}$ $\Rightarrow$ $\{g\}$.

Because larger rules are derived from smaller rules via left- or right-expansions, it is necessary to clarify whether an item can expand left or right. Next, the relevant definitions are provided.

\begin{definition}
	\label{def_15}
	\rm Assume that a sequential rule $r$: $X \Rightarrow Y$ appears in a sequence $S_c$. An item $i$ available for expanding $r$ by left expansion must satisfy $i \prec j$, $\forall j$ $\in X$, $i \notin Y$, and $X \cup \{i\}$ $\Rightarrow Y$ appear in $S_c$. Similarly, an item $i$ capable of expanding $r$ via right expansion must fulfill the condition that $i \prec j$, $\forall j$ $\in Y$, $i \notin X$, and simultaneously $X \Rightarrow$ $Y \cup$ $\{i\} $ appear in $S_c$ \cite{zida2015efficient}. 
\end{definition}

In Definition \ref{def_15}, it is discovered that in a sequence $S_c$, some items can only expand rule $r$ through left expansion, denoted as \textit{onlyLeft}$(r, S_c)$, and some items only expand rules via right expansion expressed as \textit{onlyRight}$(r, S_c)$, while some other items have both properties represented as \textit{leftRight}$(r, S_c)$. After laying the foundation, it is time to formally define the utility-list.

\begin{definition}
	\label{def_16}
	\rm The utility-list of a rule $r$ is denoted as $UL(r)$. Each row of a utility-list consists of a five-tuple ($S_{sid}$, \textit{iutil}, \textit{lutil}, \textit{rutil}, \textit{lrutil}), where $S_{id}$ indicates the identifier of the supporting sequence, the element \textit{iutil} implies the real utility of $r$ in sequence $S_{sid}$, and the element \textit{lutil} indicates the sum of utilities of items in \textit{onlyLeft}$(r, S_{sid})$. Similarly, the elements \textit{rutil} and \textit{lrutil} are the sum of utilities from those items in \textit{onlyRight} and \textit{leftRight}, respectively.
\end{definition}

\begin{figure}[h]
	\centering
	\includegraphics[scale=0.4]{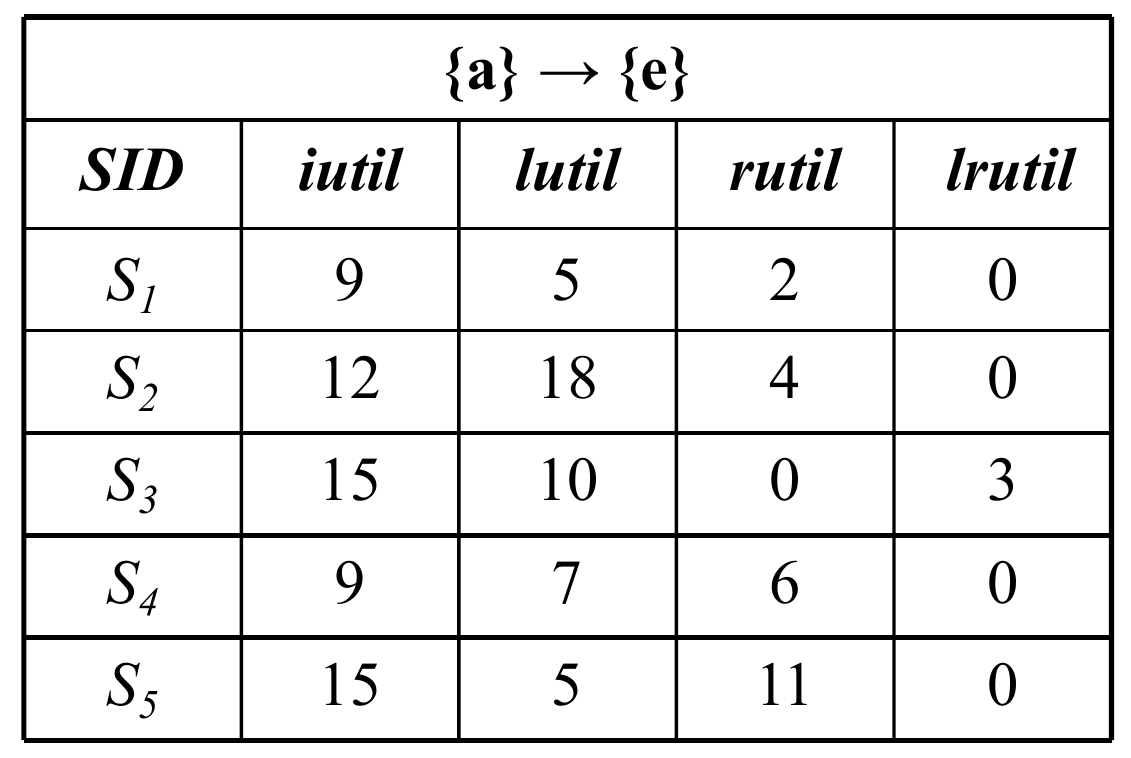}
	\caption{The utility-list of $\{a\} \Rightarrow \{e\}$}
	\label{fig:utilityTable}
\end{figure}

For example, the rule $r$: $\{a\}$ $\Rightarrow$ $\{e\}$ presented in Fig. \ref{fig:utilityTable} occurs in the sequences $S_1$, $S_2$, $S_3$, $S_4$, and $S_5$. In sequence $S_1$, the value of \textit{iutil} is equal to the sum of the utilities of $a$ and $e$. According to Definition \ref{def_16}, item $b$ belongs to \textit{onlyLeft}$(r, S_1)$, whereas \textit{leftRight}$(r, S_1)$ is an empty set. \textit{onlyRight}$(r, S_1)$ contains $g$. Therefore, the element \textit{lutil} in $S_1$ equals $u(b, S_1)$ = \$5, \textit{lrutil} = \$0, and \textit{rutil} = $u(g, S_1)$ = \$2. The elements in the other sequences can also be calculated using the same method. After constructing the initial utility-lists, it is clear that they have significant hidden properties that can be exploited.

\begin{property}
	\label{pro_3}
	\rm The utility of a rule $r$ in a sequence database is equivalent to the sum of \textit{iutil} in each row of $UL(r)$ \cite{zida2015efficient}.
\end{property}

\begin{property}
	\label{pro_4}
	\rm The support of the given rule $r$ is corresponding to the number of rows in $UL(r)$ \cite{zida2015efficient}. 
\end{property}

\begin{property}
	\label{pro_5}
	\rm An interesting observation in the utility-list $UL(r)$ corresponding to rule $r$ is that the sum of \textit{iutil}, \textit{lutil}, \textit{rutil}, and \textit{lrutil} of all tuples is always greater than $u(r)$, as are the utilities of left- and right-expansions of $r$. Moreover, because the sum is less than \textit{SEU}$(r)$ for good, it has a tighter upper bound than \textit{SEU}$(r)$ \cite{zida2015efficient}.
\end{property}

\begin{property}
	\label{pro_6}
	\rm Another observation is that the utilities of the left-expansions of a rule $r$ in $UL(r)$ are no greater than the sum of \textit{iutil}, \textit{lutil}, and \textit{lrutil}. Similarly, it has a more compact upper bound than \textit{SEU}$(r)$ \cite{zida2015efficient}.
\end{property}

With a more compact utility upper bound, the above two properties can be used in the expanding operation to reduce the search space even more.

\begin{strategy}
	\label{stra_3}
	\rm Let a sequential rule $r$ correspond to a utility-list $UL(r)$. If the sum of utilities in $UL(r)$ is less than \textit{minutil}, $r$ and its right expansions are capable of being pruned ahead of time \cite{zida2015efficient}.
\end{strategy}

\begin{strategy}
	\label{stra_4}
	\rm Let the utility-list of sequential rule $r$ be $UL(r)$. If the sum of utilities except \textit{rutil} in $UL(r)$ is less than \textit{minutil}, $r$ and its left expansions are capable of being pruned in advance \cite{zida2015efficient}.
\end{strategy}

There is no need to explore the database again to construct the utility-lists of larger rules after building the utility-lists of all 1*1 rules. We can derive the utility-lists of larger rules by taking advantage of smaller ones through pattern growth. We assume that a sequential rule $r$ is expanded with an item $i$ to acquire a new rule $r'$. A tuple of the utility-list of $r$ and $r'$ in a sequence $S_{sid}$ can be expressed as ($S_{sid}$, \textit{iutil}, \textit{lutil}, \textit{rutil}, \textit{lrutil}) and ($S_{sid}$, \textit{iutil}$'$, \textit{lutil}$'$, \textit{rutil}$'$, \textit{lrutil}$'$), respectively. Obviously, we obtain the following equations:

\begin{equation}
	\label{equ_1}
	\textit{iutil}' = \textit{iutil} + u(i, S_{sid}) \cite{zida2015efficient}. 
\end{equation}

\begin{equation}
	\label{equ_2}
	\textit{lutil}' = \textit{lutil} - \sum u(j, S_{sid}) - u(i, S_{sid}), 
\end{equation}
where $ j \notin$ \textit{onlyLeft}$(r', S_{sid})$ $\wedge j \in$ \textit{onlyLeft}$(r, S_{sid})$, $i \in$ \textit{onlyLeft}$(r, S_{sid})$ \cite{zida2015efficient}.

\begin{equation}
	\label{equ_3}
	\textit{rutil}' = \textit{rutil} - \sum u(j, S_{sid}) - u(i, S_{sid}), 
\end{equation}
where $ j \notin $\textit{onlyRight}$(r', S_{sid})$ $\wedge j \in $ \textit{onlyRight}$(r, S_{sid})$, $i \in$ \textit{onlyRight}$(r, S_{sid})$ \cite{zida2015efficient}.

\begin{equation}
	\label{equ_4}
	\textit{lrutil}' = \textit{lrutil} - \sum u(j, S_{sid}) - u(i, S_{sid}), 
\end{equation}
where $ j \notin$ \textit{leftRight}$(r', S_{sid})$ $\wedge j \in$ \textit{leftRight}$(r, S_{sid})$, $i \in$ \textit{leftRight}$(r, S_{sid})$ \cite{zida2015efficient}.

For example, as shown in Fig. \ref{fig:utilityTable}, the rule $r$: $\{a\}$ $ \Rightarrow \{e\}$ can be expanded into $r'$: $\{a, c\}$ $\Rightarrow \{e\}$ if its upper bound is greater than \textit{minutil}. Therefore, in sequence $S_2$, \textit{iutil}$(r', S_2)$ = $u(r, S_2)$ + $u(c, S_2)$ = \$12 + \$4 = \$16, \textit{lutil}$(r', S_2)$ = \textit{lutil}$(r, S_2)$ - $u(b, S_2)$ - $u(c, S_2)$ = \$18 - \$5 - \$4 = \$9, \textit{rutil}$(r', S_1)$ = \textit{rutil}$(r, S_1)$ = \$4, and \textit{lrutil}$(r', S_1)$ = \textit{lrutil}$(r, S_1)$ = \$0.

\subsection{Optimization with Bit Vectors}

The preceding discussion focused on the properties and upper bounds of the utility; however, some improvements are made in other constrained measures. To calculate the confidence, bond, and lift of rule $r$: $X \Rightarrow Y$, we need to know \textit{sup}$(r)$, \textit{sup}$(X)$, \textit{sup}$(Y)$, \textit{dissup}$(X)$, and \textit{dissup}$(Y)$. In fact, \textit{sup}$(r)$ is easily accessible from the utility-list of $r$, where \textit{sup}$(r)$ is the number of rows in the utility-list. However, the others are difficult to calculate because it is not feasible to scan the database every time they are needed.

A practical and efficient approach is to create bit vectors for any items existing in the sequences at the beginning. Assuming a single item appears in the $j$th sequence, the $j$th bit is set to one; otherwise, it is set to zero. The support of an itemset $X$ is the length of the intersection of the bit vectors of the items in $X$. By analogy, the disjunctive support of $X$ is equal to the length of the union of the bit vectors of the items in it. For example, $bv(a)$ = 11111 and $bv(c)$ = 01001. Note that \textit{sup}$(ac)$ = $|bv(a)$ $\cap$ $bv(c)|$ = $|$01001$|$ = 2 and \textit{dissup}$(ac)$ = $|bv(a) \cup bv(c)|$ = $|$11111$|$ = 5.

This study follows the rule of allowing left expansions after right expansions but not right expansions after left expansions. If an alternative order of expansion is adopted, which allows right expansions after left expansions but not left expansions after right expansions, a critical property in terms of confidence can be discovered. The performance of the developed algorithm can be significantly improved by employing this property for pruning unqualified rules.

\begin{property}
	\label{pro_7}
	\rm If the confidence of a rule is less than \textit{minconf}, there is no need to expand it via the right expansion. 
\end{property}

\begin{proof}
	Let the rules $r$: $X \Rightarrow Y$ and $r'$: $X \Rightarrow$ $Y \cup$ $\{c\}$ be present in the database. Two equations follow from the definition of confidence: \textit{conf}$(r)$ = $|$\textit{sids(r)}$|$/$|\textit{sids}(X)|$ and \textit{conf}$(r')$ = $|\textit{sids}(r')|$/$|\textit{sids}(X)|$ hold. Therefore, the inequality \textit{conf}$(r)$ $\geq$ $\textit{conf}(r')$ holds because $|\textit{sids}(r')|$ $\leq$ $|\textit{sids}(r)|$.
\end{proof}

\subsection{The Designed Pruning Strategies}

\begin{figure}[h]
	\centering
	\includegraphics[scale=0.4]{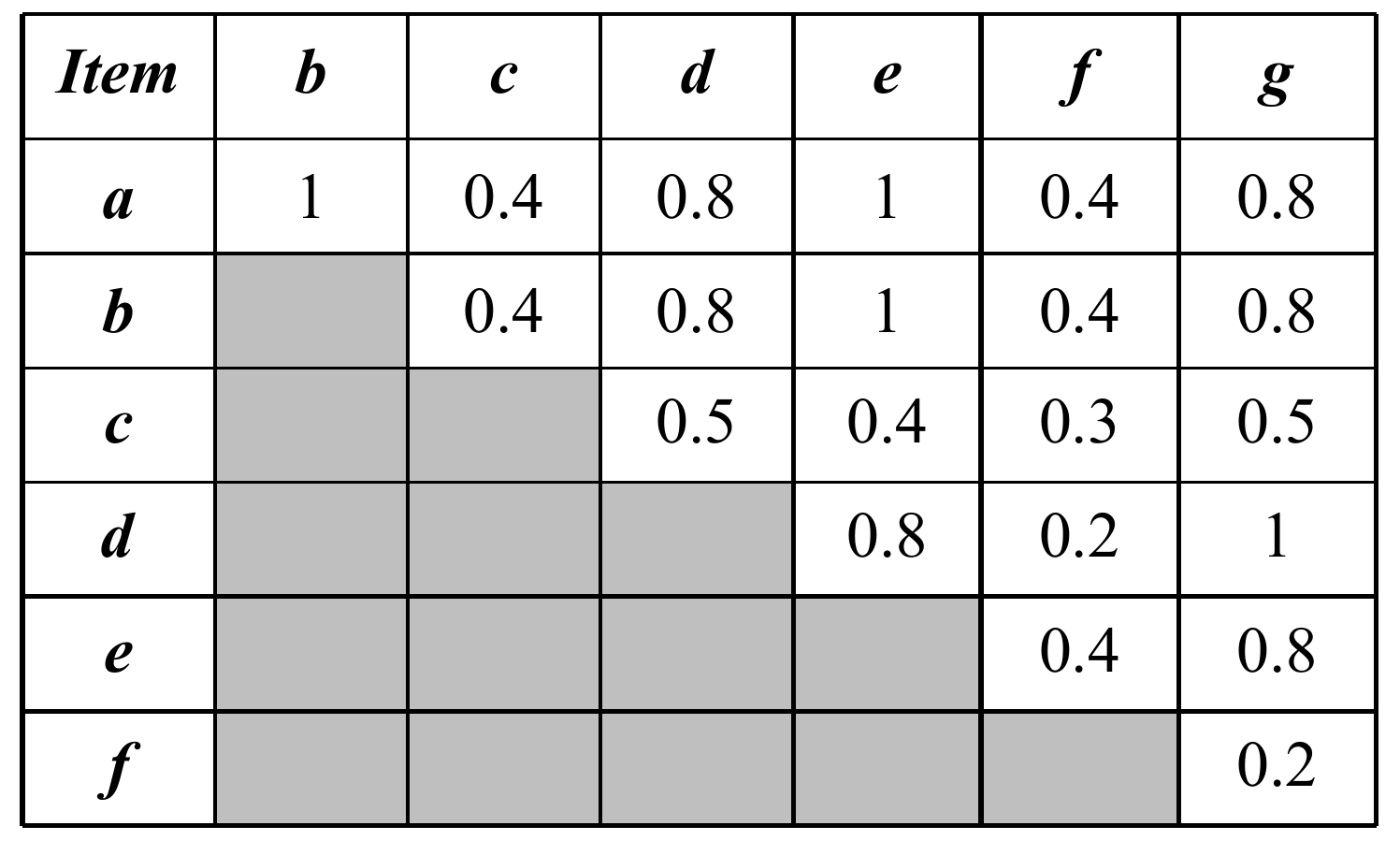}
	\caption{ The Bond Matrix}
	\label{fig:BondMatrix}
\end{figure}

In this subsection, we propose two novel data structures based on utility and bond to propose some pruning strategies and to abandon unpromising rules in advance. This can assist in improving the efficiency and performance of the proposed algorithm. The details are as follows:

Because the bond measure is anti-monotonicity, a structure similar to EUCS \cite{fournier2014fhm} is designed to maintain the bond value between two items during the second scans of the sequence database. The \textit{Bond Matrix} is developed as a series of triples of the form BondMatrix($a$, $b$) = $c$, where $a \in I$, $b \in I$, $a \prec b$ by dictionary order, and $c \in [0, 1]$. This implies that $c$ is the bond between $a$ and $b$, which is the \textit{bond}$(ab)$. For example, Fig. \ref{fig:BondMatrix} depicts the bond matrix of the items in Table \ref{table:SD}.

\begin{strategy}
	\label{stra_5}
	\rm Assume that the last items of the antecedent and consequent of a rule $r$: $X \Rightarrow Y$ are $x$ and $y$, respectively. If $r$ is left extended with an item $i$ to form $r'$: $X \cup$ $\{i\}$ $\Rightarrow Y$ such that BondMatrix($x$, $i$) = null or BondMatrix($x$, $i$) $<$ \textit{minbond}, $r'$ and all its expansions can be directly pruned. Similarly, if $r$ is right extended with an item $i$ to form $r'$: $X$ $\Rightarrow$ $Y \cup \{i\}$ such that BondMatrix($y$, $i$) = null or BondMatrix($y$, $i$) $<$ \textit{minbond}, $r'$ and all its expansions can also be directly pruned.
\end{strategy}

\begin{figure}[h]
	\centering
	\includegraphics[scale=0.4]{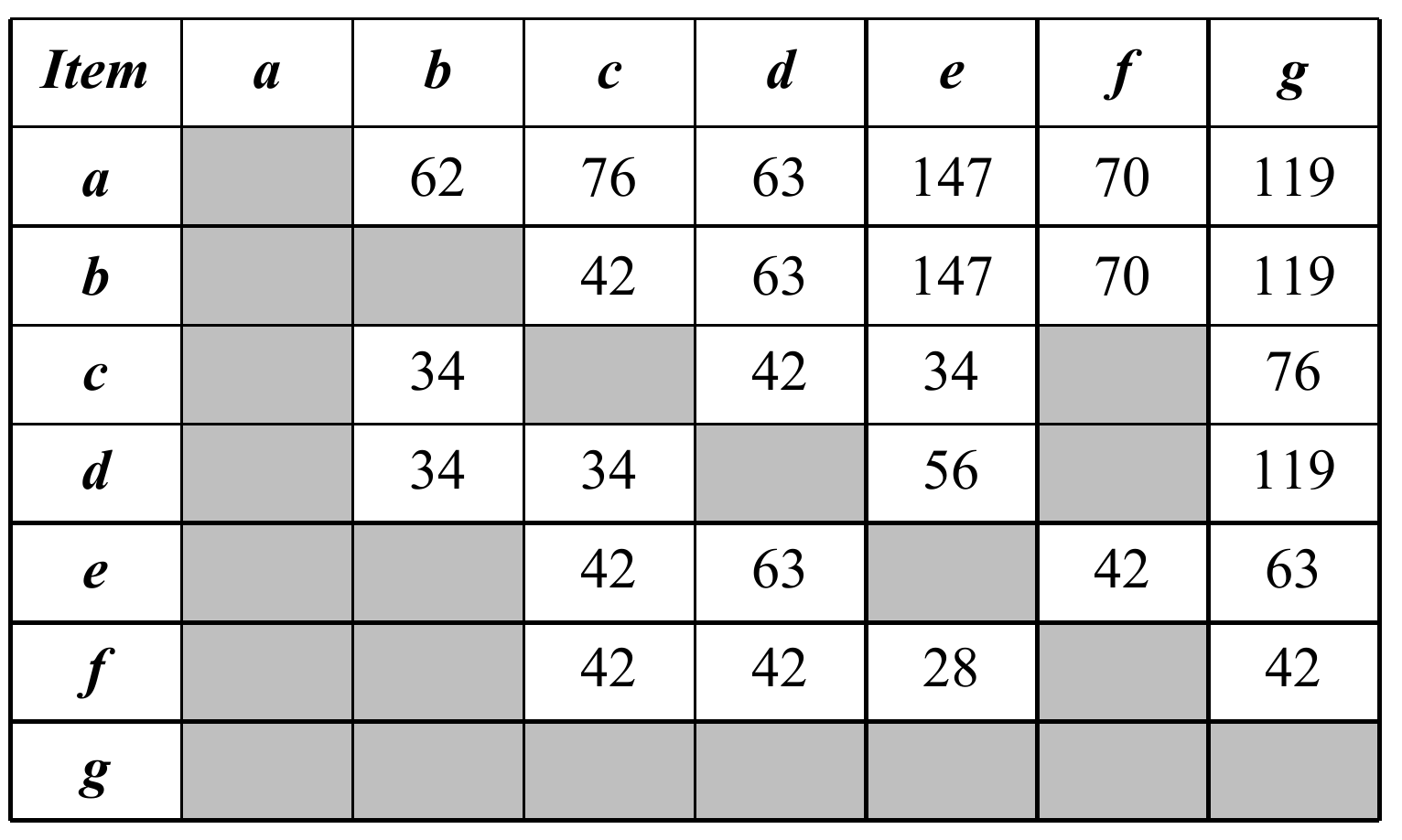}
	\caption{The ESUCS structure}
	\label{fig:ESUCS}
\end{figure}

Because the sequence-estimated utility of a rule has a downward closure property, we design a \textit{estimated sequence utility co-occurrence structure} (ESUCS) to maintain the \textit{SEU} between two items. The ESUCS is a set of triples of the form ESUCS($a$, $b$) = $c$, where $a \in I$, $b \in I$, and $c \in R^+$. Note that $a$ and $b$ belong to different itemsets, with $a$ appearing first in temporal order and $b$ appearing later. For example, the ESUCS formed by a running example is illustrated in Fig. \ref{fig:ESUCS}. ESUCS is similar to EUCS \cite{fournier2014fhm} except that with the addition of the order between items, the values of ESUCS ($a$, $b$) and ESUCS ($b$, $a$) are not identical, and they represent the \textit{SEU} of the rules $a \Rightarrow b$ and $b \Rightarrow a$, respectively. While there is no order between items in EUCS, EUCS ($a$, $b$) is equal to EUCS ($b$, $a$), which is an upper triangular structure with symmetry.

\begin{strategy}
	\label{stra_6}
	\rm Assume that the last items of the antecedent and consequent of a rule $r$: $X \Rightarrow Y$ are $x$ and $y$, respectively. If $r$ is left extended with an item $i$ to form $r'$: $X \cup$ $\{i\}$ $\Rightarrow Y$; thus, ESUCS($i$, $y$) = null or ESUCS($i$, $y$) $<$ \textit{minutil}, then $r'$ and all its expansions can be immediately trimmed. Similarly, if $r$ is right extended with an item $i$ to form $r'$: $X \Rightarrow Y$ $\cup \{i\}$, resulting in ESUCS($x$, $i$) = null or ESUCS($x$, $i$) $<$ \textit{minutil}, then $r'$ and all its expansions can also be immediately trimmed.
\end{strategy}

\subsection{The Proposed Algorithm}

Algorithm \ref{CoUSR} exhibits the core pseudocode of the CoUSR, integrating the critical aspects from previous discussions. The CoUSR algorithm has five inputs, and they are a sequence database \textit{SD}, the minimum utility threshold \textit{minutil}, the minimum confidence threshold \textit{minconf}, the minimum bond threshold \textit{minbond}, and the minimum lift threshold \textit{minlift}. The algorithm assumes complete CHUSRs.

\begin{algorithm}[h]
	\label{CoUSR}
	\caption{The CoUSR algorithm}
	\LinesNumbered  
	\KwIn{\textit{SD}, a sequence database; \textit{minutil}, the minimum utility thereshold; \textit{minconf}, the minimum confidence thereshold; \textit{minbond}, the minimum bond thereshold; \textit{minlift}, the minimum lift thereshold.} 
	\KwOut{a collection of CHUSRs.} 
	
	scan \textit{SD} to calculate the \textit{SEU} of each item $i \in I$\; 
	$I^*$ $\leftarrow$ $\{i|i \in I \wedge \textit{SEU}(i)$ $\geq \textit{minutil}\}$\; 
	 filter out items from \textit{SD} that $j \in I$ such that $j \notin I^*$\;
	scan \textit{SD} to calculate the bit vector of each item $i \in I^*$ and build the bond matrix, BondMatrix\;
	scan \textit{SD} to calculate $R$, the set of rules of the form $r$ : $i \Rightarrow j (i, j \in I^*) $ appearing in \textit{SD} and calculate \textit{SEU(r)}, \textit{sids(r)} and meanwhile build ESUCS\;
	$R^*$ $\leftarrow$ $\{r|r \in R \cap \textit{SEU}(r)$ $\geq \textit{minutil}\}$\; 
	construct the initial utility-lists $UL(r)$,  $r$ $ \in R^*$\;
	
	\For{each rule $r \in$ $R^*$}{
		\If{$UL(r)$.\textit{iutil} $\geq$ \textit{minutil} $\wedge$ \textit{conf}$(r) \geq$ \textit{minconf} $\wedge$ \textit{lift}$(r) \geq$ \textit{minlift}}{
			\textit{CHUSR} $\leftarrow$ \textit{CHUSR} $ \cup$  $r$\;
		}
		
		\If{($UL(r)$.\textit{iutil} + $UL(r)$.\textit{lutil} + $UL(r)$.\textit{rutil} + $UL(r)$.\textit{lrutil}) $\geq$ \textit{minutil}}{
			call \textbf{RightExpansion}($r$, \textit{SD}, \textit{minutil}, \textit{minconf}, \textit{minbond}, \textit{minlift}, \textit{SIDsX}, \textit{SIDsY}, \textit{SIDsORX}, \textit{SIDsORY})\;
		}
		
		\If{($UL(r)$.\textit{iutil} + $UL(r)$.\textit{lutil} + $UL(r)$.\textit{lrutil}) $\geq$ \textit{minutil}}{
			call \textbf{LeftExpansion}($r$, \textit{SD}, \textit{minutil}, \textit{minconf}, \textit{minbond}, \textit{minlift}, \textit{SIDsX}, \textit{SIDsY}, \textit{SIDsORX})\;
		}
	}	
\end{algorithm}

\begin{algorithm}[h]
	\label{Right}
	\caption{RightExpansion($r$, \textit{SD}, \textit{minutil}, \textit{minconf}, \textit{minbond}, \textit{minlift}, \textit{SIDsX}, \textit{SIDsY}, \textit{SIDsORX}, \textit{SIDsORY})}
	\LinesNumbered  
	\KwIn{$r$: $X \Rightarrow Y$, a sequence rule; \textit{SD};  \textit{minutil}; \textit{minconf}; \textit{minbond}; \textit{minlift}; \textit{SIDsX}, the set of sequences containing $X$; \textit{SIDsY}, the set of sequences containing $Y$; \textit{SIDsORX}, the set of sequences containing any one of the items in $X$; \textit{SIDsORY}, the set of sequences containing any one of the items in $Y$.} 
	
	\For{sequence $S \in$ \textit{sids}$(r)$}
	{
		
		\For{rule $e$: $X \Rightarrow Y$ $\cup \{i\}$ $|i \in$ \textit{leftRight}$(r, S) \cup$ \textit{onlyRight}$(r, S)$ $ \cup$ \textit{onlyLeft}$(r, S)$}
		{
			$x$ is the last item in $X$\;
			$y$ is the last item in $Y$\;
			\If{ESUCS($x$, $i$) $<$ \textit{minutil}}
			{
				continue\;
			}
			\If{BondMatrix($y$, $i$) $<$ \textit{minbond}}
			{
				continue\;
			}
			\If{\textit{bond}$(Y \cup i) \geq$ \textit{minbond}}
			{
				\textit{newSIDsY} =  \textit{SIDsY} $\cap$ \textit{sids}$(i)$\;
				\textit{newSIDsORY} =  \textit{SIDsORY} $\cup$ \textit{sids}$(i)$\;
				\textit{conf}$(e)$ = $|\textit{sids}(e)|$ / $|\textit{SIDsX}|$\;
				\textit{lift}$(e)$ = $(|SD|$ $\times$ $|\textit{sids}(e)|)$/$(|\textit{SIDsX}|$ $\times$ $|\textit{newSIDsY}|)$\;
				\If{$UL(e)$.\textit{iutil} $\geq$ \textit{minutil} $\wedge$ \textit{conf}$(e) \geq$ \textit{minconf} $\wedge$ \textit{lift}$(e) \geq$ \textit{minlift}}
				{
					\textit{CHUSR} $\leftarrow$ \textit{CHUSR} $ \cup$  $e$\;
				}
				\If{($UL(e)$.\textit{iutil} + $UL(e)$.\textit{lutil} + $UL(e)$.\textit{rutil} + $UL(e)$.\textit{lrutil}) $\geq$ \textit{minutil}}
				{
					call \textbf{RightExpansion}($e$, \textit{SD}, \textit{minutil}, \textit{minconf}, \textit{minbond}, \textit{minlift}, \textit{SIDsX}, \textit{newSIDsY}, \textit{SIDsORX}, \textit{newSIDsORY})\;
				}
				\If{($UL(e)$.\textit{iutil} + $UL(e)$.\textit{lutil} + $UL(e)$.\textit{lrutil}) $\geq$ \textit{minutil}}
				{
					call \textbf{LeftExpansion}($e$, \textit{SD}, \textit{minutil}, \textit{minconf}, \textit{minbond}, \textit{minlift}, \textit{SIDsX}, \textit{newSIDsY}, \textit{SIDsORX})\;
				}
			}
		}
	}	
	
\end{algorithm}

Initially, the database is iterated once to obtain the \textit{SEU} of all items in $I$, and filter out some unsatisfactory items whose \textit{SEU} is less than \textit{minutil} to obtain $I^*$ (according to Strategy \ref{stra_1}). The database is scanned again to record the bit vector of each item in $I^*$ and construct the BondMatrix structure for later reference according to $I^*$. Subsequently, one more database scan is performed to calculate the  \textit{SEU}$(r)$ and the set of sequences \textit{sids}$(r)$ containing the rules $r$, whose size is one $\ast$ one. Simultaneously, the ESUCS is built during this database search. It is reasonable to discard unpromising rules whose \textit{SEU} is less than \textit{minutil} and access $R^*$ (according to Strategy \ref{stra_2}). At this point, initial utility-lists $UL(r)$ for each promising sequential rule $r$ can be constructed. Subsequently, the search process is initiated. This is the depth-first search algorithm. The utility of each rule $r$ in $R*$ can be easily obtained using $UL(r)$, and the confidence and lift values can be obtained by intersecting and combining the bit vectors. If a rule $r$ simultaneously satisfies that the utility is greater than \textit{minutil}, the confidence is greater than \textit{minconf}, and the list is greater than \textit{minlift}, it is a CHUSR and stored in the \textit{CHUSR} collection. If the sum of utilities in $UL(r)$ satisfies the Strategy \ref{stra_3}, \textbf{RightExpansion} is called. At the same level, if the sum of utilities except \textit{rutil} in $UL(r)$ satisfies the Strategy \ref{stra_4}, \textbf{LeftExpansion} is called.

\begin{algorithm}[h]
	\label{Left}
	\caption{LeftExpansion($r$, \textit{SD}, \textit{minutil}, \textit{minconf}, \textit{minbond}, \textit{minlift}, \textit{SIDsX}, \textit{SIDsY}, \textit{SIDsORX})}
	\LinesNumbered 
	\KwIn{$r$: $X \Rightarrow Y$; \textit{SD};  \textit{minutil}; \textit{minconf}; \textit{minbond}; \textit{minlift}; \textit{SIDsX};  \textit{SIDsY};  \textit{SIDsORX}.} 
	
	\For{\rm sequence $S \in$ \textit{sids}$(r)$}
	{
		
		\For{\rm rule $e$: $X \cup \{i\}$ $\Rightarrow$ $Y$ $ |i \in$ \textit{leftRight}$(r, S) \cup$ \textit{onlyLeft}$(r, S)$}
		{
			$x$ is the last item in $X$\;
			$y$ is the last item in $Y$\;
			\If{ESUCS($i$, $y$) $<$ \textit{minutil}}
			{
				continue\;
			}
			\If{BondMatrix($x$, $i$) $<$ \textit{minbond}}
			{
				continue\;
			}
			\If{\textit{bond}$(X \cup i)$ $\geq$ \textit{minbond}}
			{
				\textit{newSIDsX} =  \textit{SIDsX} $\cap$ \textit{sids}$(i)$\;
				\textit{newSIDsORX} =  \textit{SIDsORX} $\cup$ \textit{sids}$(i)$\;
				\textit{conf}$(e)$ = $|\textit{sids}(e)|$ / $|\textit{newSIDsX}|$\;
				\textit{lift}$(e)$ = $(|SD|$ $\times$ $ |\textit{sids}(e)|)$/$(|\textit{newSIDsX}|$ $\times $ $ |\textit{SIDsY}|)$\;
				\If{$UL(e)$.\textit{iutil} $\geq$ \textit{minutil} $\wedge$ \textit{conf}$(e)$ $\geq$ \textit{minconf} $\wedge$ \textit{lift}$(e)$ $\geq$ \textit{minlift}}
				{
					\textit{CHUSR} $\leftarrow$ \textit{CHUSR} $ \cup$  $e$\;
				}
				\If{$UL(e)$.\textit{iutil} + $UL(e)$.\textit{lutil} + $UL(e)$.\textit{lrutil}) $\geq$ \textit{minutil}}
				{
					call \textbf{LeftExpansion}($e$, \textit{SD}, \textit{minutil}, \textit{minconf}, \textit{minbond}, \textit{minlift}, \textit{newSIDsX}, \textit{SIDsY}, \textit{newSIDsORX})\;
				}
			}
		}
	}
	
\end{algorithm}

Algorithm \ref{Right} provides details of the right expansion. The RightExpansion function takes the rules to be extended $r$: $X \Rightarrow Y$, a sequence database \textit{SD}, the four restraint thresholds \textit{minutil}, \textit{minconf}, \textit{minbond}, and \textit{minlift}, the collection of sequences containing $X$ \textit{SIDsX}, the collection of sequences containing $Y$ \textit{SIDsY}, the collection of sequences containing any item in $X$ \textit{SIDsORX}, and the collection of sequences containing any item in $Y$ \textit{SIDsORY} as input. Let  $r$ expand with an item $i$ in \textit{onlyLeft}$(r, s)$, \textit{leftRight}$(r, s)$ or \textit{onlyRight}$(r, s)$ by right, $x$ is the last item in $X$ that is the greatest item, and $y$ is the last item in $Y$. If ESUCS($x$, $i$) $<$ \textit{minutil} holds, then the rule and its expansions cannot be CHUSRs (according to Strategy \ref{stra_6}). If BondMatrix($y$, $i$) $<$ \textit{minbond} holds, it can also be ignored (according to Strategy \ref{stra_5}). If \textit{bond}$(Y \cup i) \geq$ \textit{minbond} holds, the extended rule $e$ is a promising rule (according to Strategy \ref{stra_bond}). If the rule simultaneously satisfies the utility, confidence, and lift constraints, it is added to the \textit{CHUSR} collection. As described above, the rule is expanded when the right or left expansion property is satisfied. The LeftExpansion procedure depicted in Algorithm \ref{Left} is similar to the RightExpansion procedure. However, the main difference is that the right expansion cannot be performed after the left expansion.

\section{Experiments}  \label{sec:experiments}

To evaluate the effectiveness and efficiency of the proposed algorithm and strategies, we conduct experiments in terms of runtime and memory consumption, the number of generated sequential rules, and pruned rules. Note that this is the first study to discover CHUSRs and that the benchmark algorithm is HUSRM. Although the CoUSR algorithm is proposed on the basis of the HUSRM algorithm, we do not intend to use HUSRM as a comparison because CoUSR has added two parameters that are not comparable: \textit{minbond} and \textit{minlift}. Additionally, when \textit{minbond} and \textit{minlift} are both fixed at zero, the number of rules derived is the same as that of the HUSRM. However, the runtime of HUSRM is often much longer than that of CoUSR when the same \textit{minutil} and \textit{minconf} are used, but \textit{minbond} and \textit{minlift} differ by tens of thousands of seconds in some cases. Therefore, we do not compare them here.

The efficiency of Strategies \ref{stra_1}, \ref{stra_2} \ref{stra_bond},  \ref{stra_3}, and \ref{stra_4} has been verified in this study with respect to the HUSRM algorithm \cite{zida2015efficient} and FCHM \cite{fournier2016mining}, and it is not repeated here. In this study, we compare Strategies \ref{stra_5} and \ref{stra_6}, focusing on four versions of the CoUSR algorithm (CoUSR, CoUSR$_{\_6}$, CoUSR$_{\_7}$, CoUSR$_{\_6\_7}$). CoUSR denotes the designed algorithm without Strategies \ref{stra_5} and \ref{stra_6}. CoUSR$_{\_6}$ and CoUSR$_{\_7}$ imply CoUSR with Strategies \ref{stra_5} and \ref{stra_6}, respectively, whereas CoUSR$_{\_6\_7}$ incorporates both strategies. For the convenience of plotting and illustration, we denote $\alpha$ for \textit{minutil}, $\beta$ for \textit{minconf}, $\gamma$ for \textit{minbond}, and $\delta$ for \textit{minlift}. Because there are four restriction parameters in this algorithm, to comprehensively analyze the algorithm, we mainly consider the variations in utility as representatives to evaluate the performance of the algorithm.

\subsection{Experimental Environment and Datasets}

All codes are implemented in the Java language. The experiments were performed on a PC with an Intel(R) Core(TM) i5-8500 CPU @3.00 GHz and 24 GB of RAM, which uses the Windows 10 operating system. Six realistic datasets with different characteristics were applied to ensure the superiority of the proposed algorithm from all aspects (sparse and dense). These algorithms are Bible, BMS, Kosarak10k, Sign, Scalability10K, and Yoochoose, and the specific descriptions of their features can be found in Table \ref{table:Characters}. $|D|$ and $|I|$ represent the number of sequences and distinct items in a database, respectively. \textit{avg(D)} and \textit{avg(I)} imply the average length of sequences and itemsets, respectively.

\begin{table}[h]
	\centering
	\small
	\caption{Dataset Characters}
	\label{table:Characters}
	\begin{tabular}{|c|c|c|c|c|}
		\hline
		\textbf{Dataset}    &  \textbf{$|D|$}  &  \textbf{$|I|$}  &   \textbf{\textit{avg(D)}}  &   \textbf{\textit{avg(I)}}     \\ \hline 
		Bible &   36,369   &   13,905  & 21.64  & 1 \\ \hline
		Sign &  730  &  267  &  51.99 & 1  \\ \hline
		Kosarak10k &  10,000  &  10,094  &  8.14 & 1   \\ \hline
		BMS &  59,601  &  3,340  &  4.62 & 1  \\ \hline
		Scalability10K &  1,000  &  7,312  &  6.22 &  4.35  \\ \hline
		Yoochoose &  234,300  &  16,004  &  1.13  & 1.97 \\ \hline
	\end{tabular}
\end{table}

\subsection{Runtime Comparison}

\begin{figure*}[ht]
	\centering
	\includegraphics[trim=100 0 30 0, height=0.35\textheight, width=1.05\linewidth]{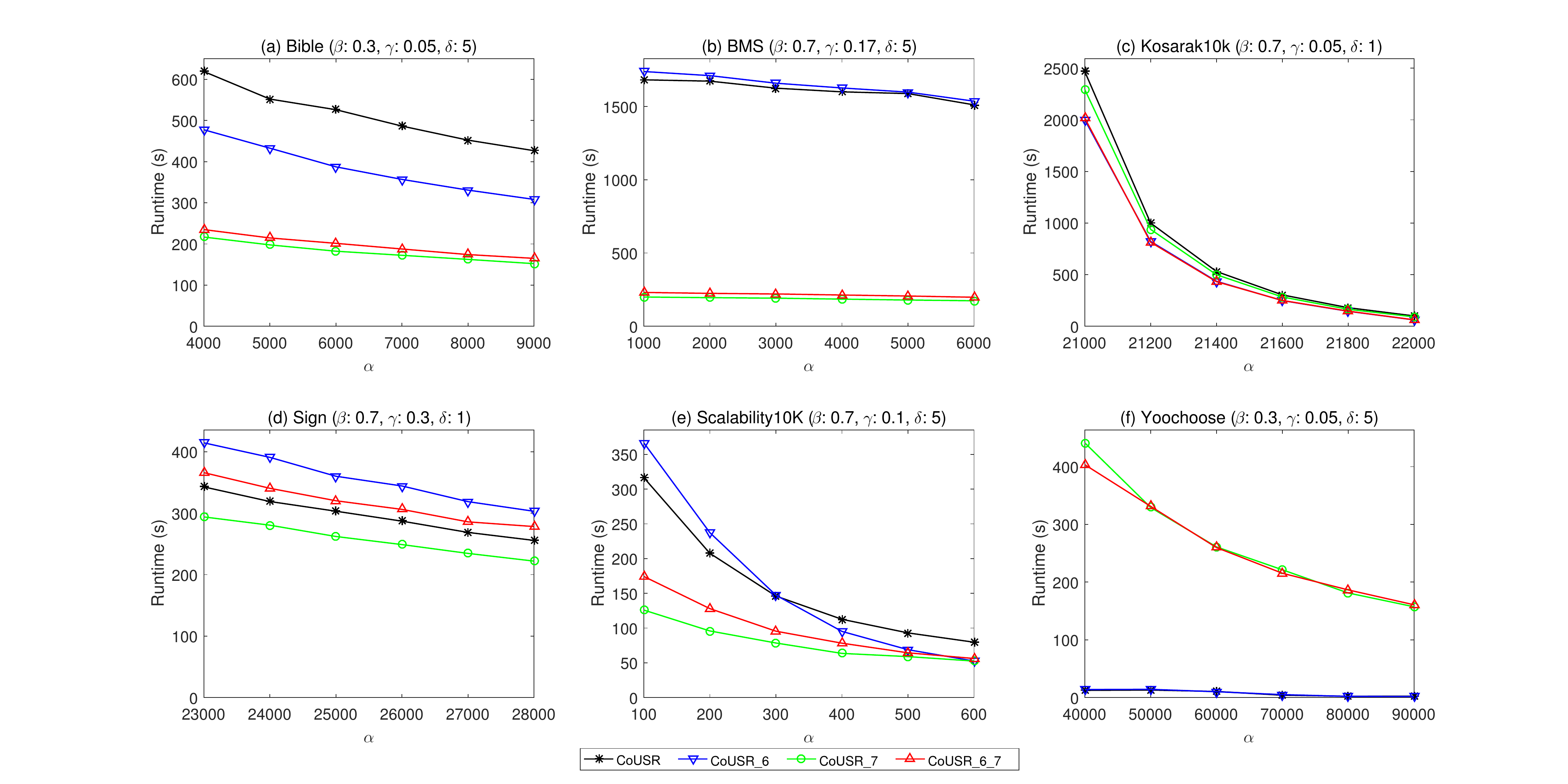}
	\caption{ Runtime with varying \textit{minutil}.}
	\label{fig:CoUSR_runtime}
\end{figure*}

As is well known, the runtime of algorithms is an important aspect of effectiveness analysis. Therefore, we compare the execution times of the methods under the assumption that $\alpha$ varies, while the other parameters are fixed. Because each dataset has different characteristics, the corresponding parameters differ for each dataset. We obtain the variation in runtime in different contexts when $\alpha$ evolves, as shown in Fig. \ref{fig:CoUSR_runtime}. It is easy to notice that as $\alpha$ increases, the execution time of the progression decreases, as do the gaps between them. Overall, CoUSR takes the longest time without the two pruning strategies, followed by CoUSR$_{\_6}$ and CoUSR$_{\_7}$, and CoUSR$_{\_6\_7}$ takes the shortest running time for both strategies. Additionally, note that CoUSR$_{\_7}$ takes less time than CoUSR$_{\_6}$. This demonstrates that the correlation-based pruning strategy does not outperform the utility-based pruning strategy. Moreover, note that in most cases, the operation times of CoUSR and CoUSR$_{\_6}$ are close, as are the operation times of CoUSR$_{\_7}$ and CoUSR$_{\_6\_7}$. For example, in the BMS dataset, as shown in Fig. \ref{fig:CoUSR_runtime}(b), $\beta$ is set to 0.7, $\gamma$ to 0.17, $\delta$ to 5, and $\alpha$ is varied from 1000 to 6000 with 1000 increments. The runtime of CoUSR decreases from 1682s to 1511s, while that of CoUSR$_{\_6\_7}$ decreases from 230s to 197s, which is much less than the former. When the operation times of CoUSR$_{\_6}$ and CoUSR$_{\_7}$ are compared, CoUSR$_{\_7}$ plays a more significant role.

\subsection{Memory Consumption}

\begin{figure*}[ht]
	\centering
	\includegraphics[trim=100 0 30 0, height=0.35\textheight, width=1.05\linewidth]{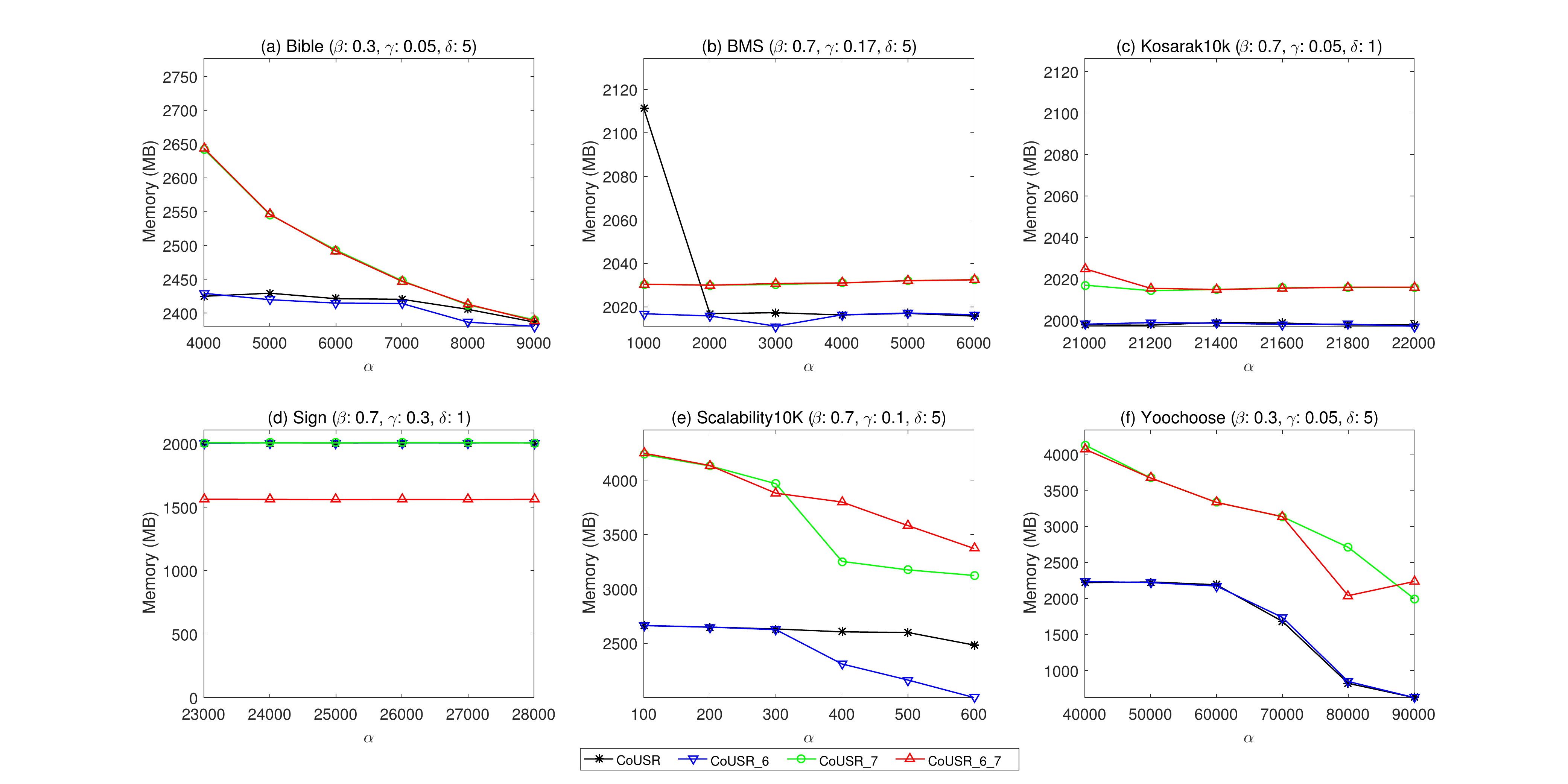}
	\caption{Memory with varying \textit{minutil}.}
	\label{fig:CoUSR_memory}
\end{figure*}

As illustrated in Fig. \ref{fig:CoUSR_memory}, CoUSR$_{\_6\_7}$ consumes the most memory because it is required to construct both ESUCS and BondMatrix. Because the number of items in the datasets is closely related to the size of these two structures, CoUSR$_{\_6\_7}$ consumes the most memory. However, in Fig. \ref{fig:CoUSR_memory}(b), CoUSR consumes more memory than CoUSR$_{\_6\_7}$ in the BMS dataset when the utility value is low, possibly because of the large number of utility-list constructions in the initial stage. When comparing the memory consumption of the CoUSR algorithm with only Strategies \ref{stra_5} and \ref{stra_6}, CoUSR$_{\_7}$ consumes significantly more memory than CoUSR$_{\_6}$, in some cases nearly twice as much. The reason for this phenomenon is that ESUCS occupies significantly more memory than BondMatrix, with the former having a size of $|I^*|^2$ and the latter having a size of $(|I^*|-1)^2$. Although CoUSR does not need to construct either structure, its memory consumption is not the lowest, with CoUSR$_{\_6}$ consuming the least. For example, as shown in Fig. \ref{fig:CoUSR_memory}(e) when $\alpha$ is low, the memory consumption of CoUSR and CoUSR$_{\_6}$ is similar; however, as $\alpha$ increases, the memory consumption of CoUSR$_{\_6}$ decreases, and the difference from CoUSR increases. The reason for this is that the pruning capabilities of CoUSR$_{\_6}$ are not outstanding when $\alpha$ is low and require the construction of the BondMatrix structure, which consumes a lot of memory space. However, as $\alpha$ increases, the number of rules that can be pruned by Strategy \ref{stra_5} increases, resulting in the construction of fewer redundant utility-lists, thereby consuming less memory than CoUSR.

\begin{table*}[ht]
	\centering
	\caption{Pruning rules number with varying \textit{minutil}}
	\label{table:CoUSR_PruningCount}
	\begin{tabular}{|c|llllll|}
		\hline 		
		\hline		
		
		\textbf{Bible} & 4,000 & 5,000  & 6,000 & 7,000  & 8,000 & 9,000   \\ \hline
		CoUSR$_{\_6}$ & 119,425,679 & 115,863,502 & 111,955,889 & 107,574,049 & 	103,017,057 & 98,966,669 \\ \hline
		CoUSR$_{\_7}$ & 317,832,011 & 288,298,290 & 263,979,229 & 	243,325,432 & 225,761,570 & 210,762,251 \\ \hline
		CoUSR$_{\_6\_7}$ & 320,028,765 & 290,577,799 & 266,336,522 & 	245,750,140 & 228,236,503 & 213,267,557\\ \hline		
		\hline		
		
		\textbf{BMS} & 1,000 &	2,000 &	3,000 &	4,000  &	5,000 &	6,000 \\ \hline
		CoUSR$_{\_6}$ & 126,983 & 291,731 & 367,902 & 2,427,132 & 4,987,798 & 	5,847,064 \\ \hline
		CoUSR$_{\_7}$ & 639,670,352 & 631,128,769 & 615,083,071 & 	600,086,025 & 586,038,313 & 574,104,943\\ \hline
		CoUSR$_{\_6\_7}$ & 639,670,927 & 631,129,369 & 615,083,532 & 	600,099,905 & 586,068,573 & 574,135,650\\ \hline 		
		\hline
		
		\textbf{Kosarak10k}  & 21,000 & 21,200  & 21,400 & 21,600  & 21,800 & 22,000 \\ \hline
		CoUSR$_{\_6}$ & 834,697,317 & 339,284,642 & 179,873,521 & 	102,417,155 & 61,701,758 & 56,130,360\\ \hline
		CoUSR$_{\_7}$ & 439,150,757 & 183,304,554 & 99,371,882 & 58,762,678 & 37,404,977 & 23,615,658\\ \hline
		CoUSR$_{\_6\_7}$ & 837,605,098 & 342,069,760 & 182,541,259 & 	105,004,783 & 64,209,099 & 	58,519,588\\ \hline		
		\hline
		
		\textbf{Sign} & 23,000 & 24,000  & 25,000 & 26,000  & 27,000 & 28,000   \\ \hline
		CoUSR$_{\_6}$ & 73,372,239 & 70,781,542 & 69,410,641 & 65,179,889 & 	60,999,879 & 61,401,908\\ \hline
		CoUSR$_{\_7}$ & 813,582,223 & 756,826,592 & 700,969,676 & 	652,646,596 & 606,221,986 & 569,154,898\\ \hline
		CoUSR$_{\_6\_7}$ & 816,227,798 & 759,384,140 & 703,545,572 & 	655,073,816 & 608,539,612 & 571,474,179\\ \hline		
		\hline		
		
		\textbf{Scalability10K} & 100 & 200  & 300 & 400  & 500 & 600   \\ \hline 
		CoUSR$_{\_6}$ & 465,313 & 11,051,082 & 42,259,333 & 66,508,272 & 	73,546,959 & 75,581,955\\ \hline
		CoUSR$_{\_7}$ & 523,857,749 & 342,171,590 & 232,033,718 & 	172,937,045 & 139,872,875 & 117,868,255\\ \hline
		CoUSR$_{\_6\_7}$ & 523,868,528 & 342,310,284 & 232,335,631 & 	173,293,302 & 140,192,238 & 118,134,902\\ \hline		
		\hline
				
		\textbf{Yoochoose} & 4,000 & 5,000  & 6,000 & 7,000  & 8,000 & 9,000   \\ \hline
		CoUSR$_{\_6}$ & 4,748 & 4,072 & 3,389 & 3,030 & 2,633 & 2,166\\ \hline
		CoUSR$_{\_7}$ & 480,310 & 461,787 & 263,237 & 56,161 & 8,957 & 4,591\\ \hline
		CoUSR$_{\_6\_7}$ & 481,479 & 462,934 & 264,242 & 57,167 & 9,805 & 5,354\\ \hline
		
		\hline
	\end{tabular}
\end{table*}

\begin{table}[t]
	\fontsize{5.0pt}{9pt}\selectfont
	\centering
	\caption{Derived rules number with varying \textit{minutil}}
	
	\label{table:CoUSR_pattern}
	\begin{tabular}{|c|llllll|}
		\hline		\hline
		
		\textbf{Bible} ($\beta$: 0.3, $\gamma$: 0.05, $\delta$: 5)  & 4,000 & 5,000  & 6,000 & 7,000  & 8,000 & 9,000   \\ \hline
		CoUSR & 166 & 87	& 57 &	46 & 41 &39
		\\ \hline 		
		\hline
		
		\textbf{BMS} ($\beta$: 0.7, $\gamma$: 0.17, $\delta$: 5)  & 1,000 &	2,000 &	3,000 &	4,000  &	5,000 &	6,000 \\ \hline
		CoUSR & 350,344 &	165,200 & 28,752 & 2,363 & 27 & 2
		\\ \hline 		
		\hline
		
		\textbf{Kosarak10k} ($\beta$: 0.7, $\gamma$: 0.05, $\delta$: 1)  & 21,000 & 21,200  & 21,400 & 21,600  & 21,800 & 22,000   \\ \hline
		CoUSR & 6 & 6	& 6 & 6 & 6 & 6
		\\ \hline 
		
		\hline
		
		\textbf{Sign} ($\beta$: 0.7, $\gamma$: 0.3, $\delta$: 1)  & 23,000 & 24,000  & 25,000 & 26,000  & 27,000 & 28,000   \\ \hline
		CoUSR & 120 & 80 & 42 & 21 & 11 & 5
		\\ \hline 		
		\hline		
		
		\textbf{Scalability10K} ($\beta$: 0.7, $\gamma$: 0.1, $\delta$: 5)  & 100 & 200  & 300 & 400  & 500 & 600   \\ \hline
		CoUSR & 489,586 & 358,828 & 257,363 & 189,475 & 139,561 & 101,569
		\\ \hline 		
		\hline
		
		\textbf{Yoochoose} ($\beta$: 0.3, $\gamma$: 0.05, $\delta$: 5)  & 4,000 & 5,000  & 6,000 & 7,000  & 8,000 & 9,000   \\ \hline
		CoUSR & 428,233 & 373,682 & 114,426 & 4,397 & 1,057 & 32				
		
		\\		\hline
		
		\hline
	\end{tabular}
\end{table}

\subsection{Rules and Pruning Strategies Analysis}

In this subsection, we analyze the number of CoUSRs generated by the proposed algorithm with or without the designed strategies, as well as the variations of pruned rules. Table. \ref{table:CoUSR_pattern} shows that as the utility threshold increases, the number of derived rules decreases as the restrictions on the rules tighten, resulting in fewer rules being generated.

Table \ref{table:CoUSR_PruningCount} shows the variation in the number of rules pruned for different conditions, which provides a clear representation of the effectiveness of the proposed pruning strategies. Pruning rules can entail avoiding the construction of utility-lists that cannot be the CHUSRs. CoUSR$_{\_6\_7}$ eliminates the greatest number of ineligible rules because it adopts Strategies \ref{stra_5} and \ref{stra_6}. The number of pruned rules decreases with increasing utility for essentially all datasets in Table \ref{table:CoUSR_PruningCount}, which is because fewer candidate rules are generated; hence, fewer rules are pruned away. Consistent with the runtime and memory consumption results above, CoUSR$_{\_7}$ prunes out most of the non-qualified rules and dominates the pruning process. Note that the number of pruning rules in Scalability10K for CoUSR$_{\_6}$ is increasing. This is because the upper bound on the sequence-estimated utility of the items and rules already filters many candidate rules in the previous series of pruning strategies; however, as the utility increases, fewer rules are pruned by the previous strategies, allowing the number of rules pruned by the later Strategy \ref{stra_5} to increase.

\section{Conclusion and Future Studies} 
\label{sec:conclusion}

In this study, we incorporate the concept of correlation into HUSRM and propose a novel algorithm called CoUSR. We introduce two parameters, \textit{bond} and \textit{lift}, to limit the derived rules locally and globally. To consolidate important information in sequence databases and prevent multiple scans of the given database, utility-lists are proposed. Furthermore, we introduce several pruning strategies to improve the efficiency of the algorithm using the designed BondMatrix and ESUCS structures. Subsequent experiments on realistic datasets show that these strategies are indeed functional, and that the proposed algorithm performs well. In the future, we intend to broaden the application environment of this algorithm to include uncertain data or dynamic circumstances. Furthermore, owing to the problem of difficult threshold selection, it is prudent to design a top-$k$ CoUSR algorithm. Of course, performance improvement is another critical consideration.

\ifCLASSOPTIONcaptionsoff
  \newpage
\fi


\bibliographystyle{IEEEtran}
\bibliography{CoUSR}

\end{document}